%% file: iclr2020_conference.tex
\documentclass{article} 
\usepackage{iclr2020_conference,times}

\input{math_commands.tex}

\usepackage{hyperref}
\usepackage{url}
\usepackage{booktabs}
\usepackage{natbib}
\usepackage{graphicx}
\usepackage{amsmath}
\usepackage{amssymb}
\graphicspath{{./images/}}

\title{DeepHashing using Triplet Loss}

\author{Jithin James\\
Department of Computer Science\\
Govt. Model Engineering College\\
\texttt{jithinjames@mec.ac.in}
}

\iclrfinalcopy 
\begin{document}

\maketitle

\begin{abstract}
Hashing is one of the most efficient techniques for approximate nearest neighbor search for large scale image retrieval. Most of the techniques are based on hand-engineered features and do not give optimal results all the time. Deep Convolutional Neural Networks have proven to generate very effective representation of images that are used for various computer vision tasks and inspired by this there have been several Deep Hashing models like \cite{deephashing} have been proposed. These models train on the triplet loss function which can be used to train models with a superior representation capabilities. Taking the latest advancements in training using the triplet loss I propose new techniques that help the Deep Hashing models train more faster and efficiently. Experiment result \footnote{The source code is available in \href{https://github.com/jjmachan/DeepHash}{Github}} show that using the more efficient techniques for training on the triplet loss, we have obtained a 5\% percent improvement in our model compared to the original work of \cite{deephashing}. Using a larger model and more training data we can drastically improve the performance using the techniques we propose.
\end{abstract}

\section{Introduction}
Deep learning has been solving a lot of really hard problems. Its been used in a growing number of fields and every fields where it is applied, the deep learning models perform way better than the previous methods. The main advantage of a Deep, hierarchical model is that it can learn robust and effective feature sets for itself, which are more effective than their hand-engineered counter parts. 

The advent of the Internet on the other side has created a large amount of image data that have to be curated and stored in a way that allows them to be effectively searched and retrieved. Hashing is one of the most popular and powerful techniques for Approximate Nearest Neighbour (ANN) search due to its computational
and storage efficiencies. Hashing aims to map high dimensional image features
into compact hash codes or binary codes so that the Hamming distance between
hash codes approximates the Euclidean distance between image features. Previous methods have been using hand-generated features for converting the images to hashes and to store it in the database. Then using the ANN technique we can retrieve images similar to the query image from the database.

So using Deep Convolutional Neural Networks we can train models to be better hash functions that learn complex features and relationships about the images and are able to translate that into the most effective vector representations. In this work we try to show that in-fact DCNNs trained on triplet loss function using some special tricks become very effective at this.

\section{Related Work}
\label{Related_work}
This is work is mainly based on the works of \cite{deephashing} and \cite{dpsh} who  propose the idea of training a CNN to generate the binary encoding for the input images. These were trained using the Triplet loss by \cite{tripletLoss} which takes an anchor image, positive image and a negative image and trying to maximise the distance between the anchor and negative while minimising the the distance between anchor and positive images. \cite{facenet} further worked on this idea and implemented in a wide scale and used it to identify faces. The major concern with the Triplet loss function is that the number of training data increases cubically but \cite{tripletDefence} showed how effective it is and also proposed a improved Triplet loss function which I have used for the experiments.

\section{Method}

\label{Eval_Meth} All the experiments and test were performed on the CIFAR-10 dataset which contains 50k training images and 10k test images. I use the K-Nearest Neighbours (KNN) model to find the accuracy of the embeddings for the images that are generated by the trained model. The KNN model is trained on the entire training set and a subset of the images from the test set is used to calculate the accuracy. We use the KNN model to retrieve similar images that is used to calculate the Mean Average Precision (mAP). Query images, taken from the test set were used. The different experiments performed are explained below.

\begin{enumerate}

\label{exp1}\item First, I compared various pretrained models like VGG by \cite{vgg} and Resnet by \cite{resnet} to evaluate their basic representation capabilities. 
The features from the last layer (just before classification) were used for evaluation. 
The accuracy was checked in accordance with the steps described above. The results are in Table \ref{tb1}.

\label{exp2}\item Next to evaluate the performance of the triplet training loss I implemented the triplete loss proposed by \cite{facenet}
and \cite{tripletDefence} and trained it on the model that was initialized with the pretrained weights of the 
ResNet-18 model. The standard evaluation methods were used and the results are in Table \ref{tb1}.

\label{exp3}\item The major caveat to training a model with the triplet loss is the very large dataset. The number of training 
examples increase cubically for the number of training images. \cite{facenet} does mention a few online and 
offline hard mining techniques but we have to explicitly find the images that will maximize the loss. 
However, inspired by the work of \cite{tripletDefence}, I implemented a variation of triplet loss that takes in batches 
of images from every class to compute the loss. The loss I used is as follows


There are 3 variations of this that I used. SemiHardNegetive takes triplets that have a loss greater than zero but less than $\alpha$ (which is the margin). HardNegetive takes the triplets that has the maximum loss in each batch and RandomNegetive method takes random triplets that have a loss greater than zero.

\label{exp4}\item The above experiments proved the image representation capabilities of a CNN trained on the triplet loss to be very good. So building on top of \cite{deephashing} work, I created a CNN the can directly generate the binary codes for the images which can later be used for image retrieval. The loss function of the model is

\begin{equation} \label{eqloss2}
\begin{split}
L =&-\sum_{m=1}^{M} (\Theta_{q_mp_m} - \Theta_{q_mn_m} - \alpha - \log(1+e^{\Theta_{q_mp_m} - \Theta_{q_mn_m} - \alpha})) \\
&+ \lambda\sum_{n=1}^{N}||\mathbf{b}_{n} - \mathbf{u}_{n}||_2^{2}
\end{split}
\end{equation}

\enlargethispage{-\baselineskip}

where $\Theta_{i,j}$ is

\begin{equation}
\Theta_{ij} = \frac{1}{2} \mathbf{b}_i ^T \mathbf{b}_j
\end{equation}

The hyperparameters used in by \cite{deephashing} where $\alpha$=16 and $\lambda$=100 but I found this combination to be non-converging. The we turned off the quantization error completely during the starting of the training and
made $\lambda$ = 10 after 15 epoch. Regarding the alpha parameter I tried new approach by slowly increasing the alpha values every 3 epochs. This gave some interesting properties as discussed in Section  \ref{results}. 
\end{enumerate}

\section{Results}
\label{results}

In the Experiment 1, I observed that the features from the last layers of the pretrained model do 
have some representation capabilities. In Figure 1, first plot we can observe that similar objects 
like automobiles and trucks, deer and horse etc that share similar properties are in-fact, clustered 
together. Also there is improvements in the representation capabilities as the size of the model increases. So larger models will have better representation capabilities.

In Experiment 2 we can see that there is a fair amount of improvement in the representation and 
classification capabilities of the model. But the training does take a long time for it to converge and is 
not optimal. Also the additional embedding layer that was added seems to have no additional benefits.

The use of a better variant of the Triplet loss that takes into account the hard mining factor also is very
effective in training the models. Both in time and performance, these models significantly improved the representation and classification capabilities of the models. SemiHardNegetive Triplet Selector shows very fast convergence when compare to the RandomNegetive Triplet Selector but Figure 2 show us that the latter seems to have better representation (t-SNE plots) though there were no significant differences in performance measured. 

\begin{figure}[h]
\begin{center}
\label{fig1}
\centerline{\includegraphics[width=20cm]{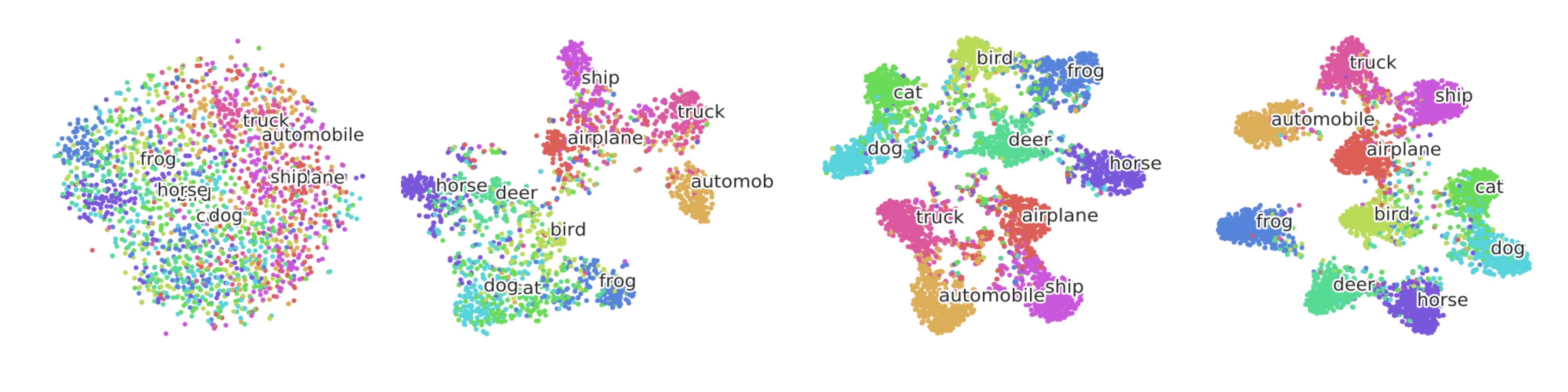}}
\caption{ Plot1 - The t-SNE plots of the last feature layer from resnet101. Plot2 - t-SNE plot after training the resnet18 model using the Triplet Loss. Plot 3,4 - t-SNE plots showing the representation of the embedding learned after the resnet18 model was trained using the SemiHard and RandomNegetive hard mining losses respectively.}
\end{center}
\end{figure}

All the results strongly supported the CNNs models capabilities of generating binary encodings that can be used for creating representation of the images in image retrieval systems. The result of the DeepHash model also support this but is was different from what was observed in the \cite{deephashing} work. The high quantization error parameter of $\lambda$ = 100 or even 10 failed to converge the model. I observed that the such a high parameter promoted the premature optimisation for the quantization error and hence the model was not able to effectively learn any useful representations. Training the model by slowly increasing the $\lambda$ parameters was a much more effective way to train and optimises both parts of the loss functions. There is a bit of representation capabilities lost due to the quantization error since the parameter $\lambda$ is low but I suspect it can be further reduced by training the model longer. In Figure 2 Plot 2 we can see that there are clusters of small dense portions but a lot of the data points are scattered about.

\begin{figure}[h]
\begin{center}
\centerline{\includegraphics[width=10cm]{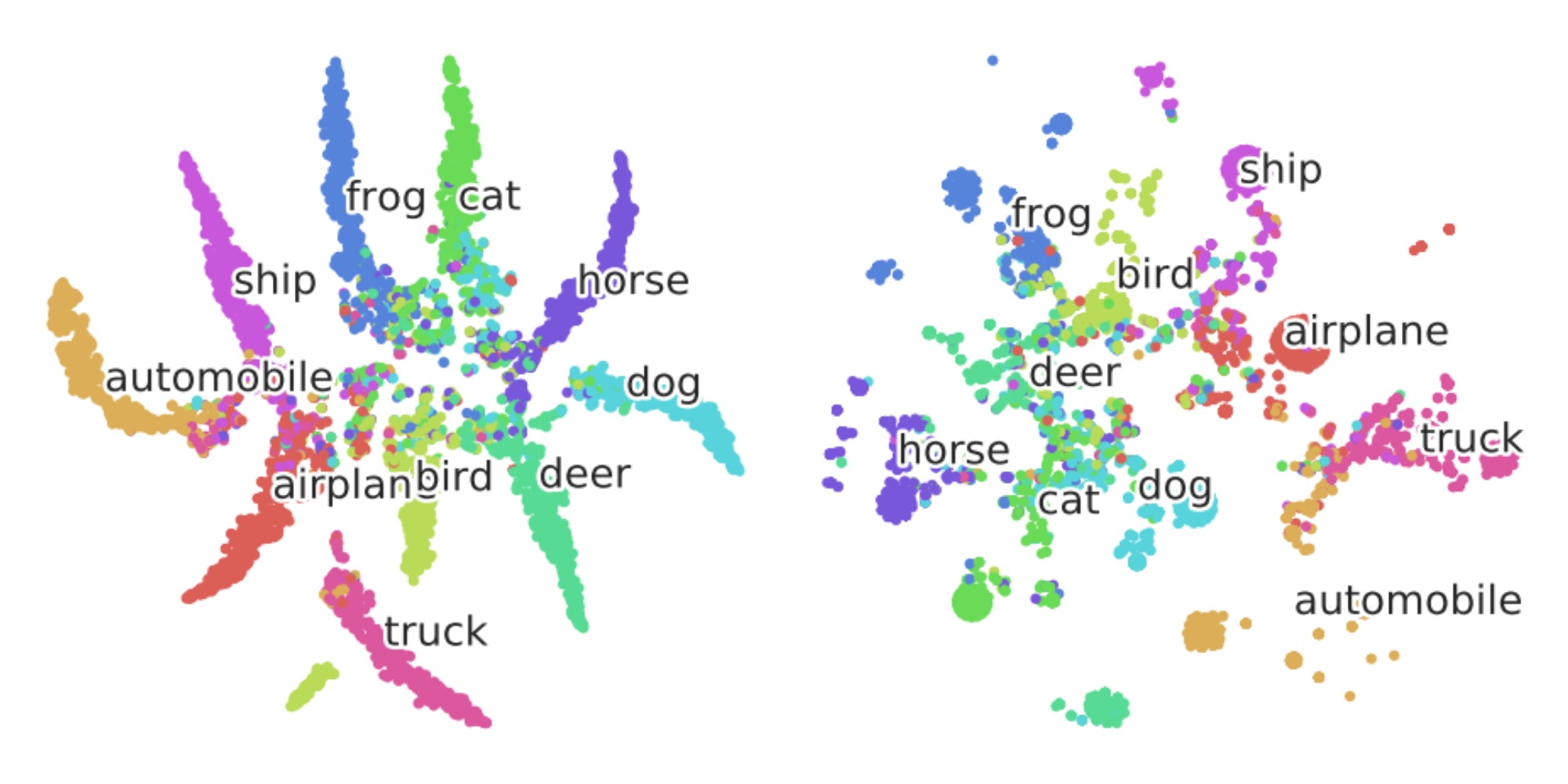}}
\end{center}
\label{fig2}
\caption{The t-SNE plot of the Hashing CNN model. The first plot is the direct output of model without applying the sign function and the second is plotted after applying the sign function.}
\end{figure}

\begin{table}[]
\label{table1}
\caption{This table shows results of Experiments 1-3. Here how the representation and retrieval capabilities of various models}
\label{tb1}
\begin{tabular}{@{}lcr@{}}
\hline
Model                                                            & KNN Accuracy & mAP    \\ \midrule
VGG16                                                            & 0.5726       & 0.4795 \\
VGG19                                                            & 0.5926       & 0.5155 \\
Resnet18                                                         & 0.5719       & 0.4619 \\
Resnet50                                                         & 0.5306       & 0.4586 \\
Resnet101                                                        & 0.5540       & 0.5171 \\ \midrule
Resnet trained on triplet loss with margin 1                     & 0.7160       & 0.6121 \\
Resnet trained on triplet loss with margin 2                     & 0.7006       & 0.6115 \\
Resnet trained on triplet loss with additional embeddings layer  & 0.6373       & 0.6168 \\ \midrule
Resnet trained using the SemiHardNegetive triplet selector       & 0.8506       & 0.8680 \\
Resnet trained using the HardestNegetive triplet selector        & 0.3653       & 0.2944 \\
Resnet trained using the RandomNegetive triplet selector         & 0.8426       & 0.8676 \\ \midrule
Resnet epoch=15  $\alpha $ = 16  $\lambda$ = 0                   & 0.7920       & \textbf{0.8483} \\
Resnet epoch=3*5  $\alpha $ = 1-16   $\lambda$ = 0 & 0.8190       & 0.7955   \\ \bottomrule
\end{tabular}
\end{table}

\section{Conclusion}
The Results show that a very effective CNN model for creating hash codes of images can be trained end-to-end using the model proposed in Experiment 4 and it shows good accuracy. Currently the Resnet18 architecture was used for the CNN model but Results how that a more Deeper model, trained longer will be able to create an even more effective hashes for the images.

\bibliography{iclr2020_conference}
\bibliographystyle{iclr2020_conference}

\end{document}

%% file: math_commands.tex
\usepackage{amsmath,amsfonts,bm}

\def\eqref#1{equation~\ref{#1}}

\def\1{\bm{1}}

\DeclareMathAlphabet{\mathsfit}{\encodingdefault}{\sfdefault}{m}{sl}
\SetMathAlphabet{\mathsfit}{bold}{\encodingdefault}{\sfdefault}{bx}{n}

